\documentclass[journal]{IEEEtran}
\usepackage{algorithm}
\usepackage{natbib}
\usepackage[noend]{algpseudocode}
\usepackage{graphicx}
\usepackage{comment}
\usepackage{tabularx}
\usepackage{array}

\usepackage{amsmath}
\usepackage{amssymb}
\begin{document}
%

\title{STARE: Predicting Decision Making Based on Spatio-Temporal Eye Movements}


%
%
%

\author{Moshe~Unger,~\thanks{Moshe Unger is with the Coller School of Management, Tel Aviv University, Israel (email: mosheunger@tauex.tau.ac.il).}
        Alexander~Tuzhilin,~\thanks{Alexander Tuzhilin is with the Stern School of Business, New York University, USA (email: atuzhili@stern.nyu.edu).}
        and~Michel~Wedel\thanks{Michel Wedel is with the Robert H. Smith School of Business, University of Maryland, USA (email: mwedel@umd.edu).}
        }

\maketitle

\begin{abstract}
The present work proposes a Deep Learning architecture for the prediction of various consumer choice behaviors from time series of raw gaze or eye fixations on images of the decision environment, for which currently no foundational models are available. The architecture, called STARE (Spatio-Temporal Attention Representation for Eye Tracking), uses a new tokenization strategy, which  involves mapping the x- and y- pixel coordinates of eye-movement time series on predefined, contiguous Regions of Interest. That tokenization makes the spatio-temporal eye-movement data available to the Chronos, a time-series foundation model based on the T5 architecture, to which co-attention and/or cross-attention is added to capture directional and/or interocular influences of eye movements. We compare STARE with several state-of-the art alternatives on multiple datasets with the purpose of predicting consumer choice behaviors from eye movements. We thus make a first step towards developing and testing DL architectures that represent visual attention dynamics rooted in the neurophysiology of eye movements.

\end{abstract}

\begin{IEEEkeywords}
eye movements, scan-path, time series, deep learning, attention mechanism, Chronos.
\end{IEEEkeywords}

%
\IEEEpeerreviewmaketitle

\section{Introduction}

\IEEEPARstart{T}he field of eye-tracking has been growing rapidly over the last two decades. 
A key problem in this field is how to 
use this data for model building purposes. There is a growing recognition in the community that eye-tracking data contains crucial information about people’s decision-making processes. 
Therefore, there has been a surge of interest in using people's eye movements while they make decisions to better understand these decision processes 
and to predict the choice outcomes \citep{martinovici2023attention, unger2024predicting} in marketing, psychology, economics, and biomedical research. For example, predicting which products consumers will pick from shelves in a supermarket based on the eye-tracking data of people exploring large collection of products on the shelves (as shown for one person by the red dots in Figure~\ref{fig:Fixations}), constitutes an important marketing problem that eye-tracking companies work on to help their customers to make product assortment and other marketing decisions. 

The traditional approach to analyzing this data consists of (a) aggregating the “raw” high-frequency eye-tracking data into sequences (called scan-paths) of eye fixations (moments when eyes are fixed on a particular point, such as when looking at an object in an image or a word when reading a text caption) and (b) using classical statistical methods to build predictive models, 
 such as when predicting the products that one chooses from the shelf. This eye-tracking paradigm has been studied over the last 20 years and has been surveyed in \cite{orquin2013attention, wedel2023modeling}. 

More recently, the traditional eye-tracking methods have been extended 
by applying advanced Deep Learning (DL) methods, such as CNNs, transformers, and metric learning, to eye-tracking analysis \cite{byrne2023predicting, polonio2015strategic, unger2024predicting}. It was shown that these advanced DL techniques can outperform the traditional  methods in predicting choice decisions. For example, Unger et al. \cite{unger2024predicting} developed a transformer-based architecture, called RETINA, that uses raw eye movement data to predict consumers' choice of one out of four options on a website. Moreover, it was shown in \cite{unger2024predicting} that the RETINA model predicted the choice better than several statistical and state-of-the-art DL methods. However, the RETINA model does not fully capitalize on the time series nature of consumers' scan-paths and was designed for data having a small number of decision alternatives (four products in this case). All this exemplifies that the field of eye-tracking is ripe for undergoing transformative changes, leading to development of the next-generation eye-tracking predictive models that would improve decision making across a wide range of applications. 

In the present research, we set out to develop a DL architecture for the time-series scan-path of eye fixations to predict choice decisions for applications having large sets of options, such as products on a shelf, as shown in Figure~\ref{fig:Fixations}. This prediction task is substantially more challenging than the more common task of predicting the choice of one out of a few options, as was done in previous eye-tracking research \cite{unger2024predicting,martinovici2023attention, shi2013information}. Our method builds on the Chronos deep learning architecture for time series \citep{ansari2024chronos} that adapts the Transformer architecture to time series data by tokenizing the data into discrete bins via scaling and quantization. Because pre-trained Chronos, despite being trained on a large corpus of time series data, does not generalize well to eye-movement data, in this paper we develop a novel tokenization approach for the x- and y- pixel coordinates of the fixation time series, and, inspired by human visual attention and neurophysiologic processes, we leverage the use of co-attention and cross-attention mechanisms to enhance predictive performance. 

Our method, called \emph{STARE} (Spatio-Temporal Attention Representation for Eye-tracking), makes it possible to annotate \emph{Regions of Interest (ROIs)} and include these predefined ROIs as tokens. While Chronos is a set of pre-trained models specifically designed for time series forecasting, capturing temporal patterns and predicting future data points in sequential data, we propose STARE to predict which products consumers choose, using data from an eye-tracking experiment provided to us by PRS in Vivo, as well as the data set from \citep{unger2024predicting}. We also predict the number of products the participants choose in the PRS-in-Vivo study, as they may choose more than one product. 

This paper makes the following contributions. First, we introduce a novel STARE model that significantly extends Chronos to the choice prediction problem. It is a gaze-driven consumer decision model that is specifically designed for the spatio-temporal eye-tracking domain to handle x- and y-coordinates of eye movements (unlike Chronos), and leverages cross-attention and co-attention mechanisms \cite{ma2019co} in two ways. In order to properly process \textit{binocular raw} eye movement data, where position information on each separate eye is available, 
cross-attention captures interocular influences (e.g., the dominant eye leading fixations), and co-attention captures separate but coordinated control of each eye. Alternatively, for \textit{eye-fixation data}, where the positions of both eyes are averaged to obtain a single fixation point, cross-attention captures directional influences (e.g., horizontal eye movement directions leading vertical ones) and co-attention captures separate but coordinated control of horizontal and vertical movements. STARE combines both attention mechanisms. We thus make a step towards developing and testing DL architectures that represent visual attention dynamics rooted in the neurophysiology of eye movements, thereby \emph{enhancing our understanding of how gaze behavior influences decision-making}. 
The plausibility of DL-based attention mechanisms to model visual attention dynamics has not been previously studied. 

Second, we propose a novel tokenization strategy for spatio-temporal eye-tracking data, making it compatible with foundational time-series models while addressing their limitations in the representation of eye-movement sequences. Traditional time-series foundation models, such as Chronos, generate tokens automatically based on time-series statistics which may not capture the structured nature of eye-movement time series. In contrast, our approach introduces decision-aware ROI tokens that segment gaze behavior into semantically meaningful regions, specifically tailored for the eye-tracking domain. By structuring gaze data in this way, our method enables more interpretable and effective modeling of \textit{in vivo} attention. Operating \emph{directly} on ROI-tokenized gaze sequences, we show that STARE significantly improves user choice prediction.

Finally, we conduct an extensive empirical evaluation using two real-world eye-tracking datasets. We compare STARE with several state-of-the-art alternatives by predicting consumer choice based on the eye-movements data. To ensure a comprehensive evaluation, we assess performance on both raw gaze trajectories and fixation-based representations, as well as for binary- and large-choice outcomes. By evaluating decision predictions across multiple classification and regression tasks, we demonstrate robustness and effectiveness of our method in capturing consumer decision-making from gaze behavior, even in data-limited settings, as shown in the experiments where we only use small fractions of the available data.

By integrating spatio-temporal tokenization, ROI-based gaze representation, and advanced attention mechanisms with a time-series foundation model, this work presents a data-driven approach for gaze-based consumer decision modeling. These contributions are at the intersection of biological and computer vision and lay the foundation for future advances in the analysis of eye-tracking data with DL for choice prediction, human-computer interaction, and cognitive inference.

\section{Background and Related Work}

\subsection{Background}

\subsubsection{Principles of Eye-Trackers and Eye Movements}
\label{overview-eyetrackers-eye-movements}

Eye movements are reliably coupled with unobserved human visual attention, 
and commonly recorded with infrared corneal reflection eye-tracking equipment. Eye-trackers work by emitting infrared light and collecting the reflections of that light off the cornea (the hard outer layer of the eye), typically at a frequency of 50Hz. 
Algorithms 
are then used to identify \emph{eye fixations} (200 to 500 msec. periods during which the eyes hardly move) from the raw eye-tracking data, averaging the positions of both eyes \cite{wedel2008eye}. Figure~\ref{fig:Fixations} illustrates the fixation points (as red circles) of a person examining a digital shelf, recorded using an eye tracker.

Time series of the x- and y- pixel coordinates of these eye fixations on images of the decision environment, such as the digital shelf, are called \emph{scan-paths}. 
To analyze the data, researchers annotate regions with semantically meaningful information on these images, which are called \emph{Regions of Interest (ROIs)}. 
For example, a few selected ROIs in Figure~\ref{fig:Fixations} are shown as yellow rectangles. 

To predict which products consumers choose from the shelf, based on eye tracking data, recently researchers have used deep-learning models, such as CNNs \cite{polonio2015strategic}. For example, Unger et al. \cite{unger2024predicting} developed a transformer-based architecture RETINA that uses raw eye movement data (at a 50 Hz frequency) to predict consumers' choice of one out of four options on a website. While that model predicted the choice better than several state-of-the-art deep learning methods, it did not fully capitalize on the time series nature of consumers' scan-paths and was designed for data having a small number of alternatives. 

The oculomotor system plays an essential role in coordinating the eye movements. In particular, eye movements are controlled by three pairs of opposing muscles attached to each eye, which control respectively horizontal, vertical and torsional eye movements \cite{wedel2008eye}. 
There are many different types of eye movements that have different functions, 
and that require highly coordinated control of the eye muscles, both for movements of the eyes in various directions in the visual field and for movements the left and right eyes. Multiple neural centers in the brain stem generate these eye movements, with sub-regions controlling different directions and types of movement \citep{sparks2002brainstem}. 

\paragraph{Horizontal and vertical eye movements} Separate neural pathways that originate in the brain stem control horizontal and vertical eye movements \citep{sparks2002brainstem}, which occur for example during systematic search where people scan search displays systematically from left-to-right and top-to-bottom. 
Horizontal eye movements 
have a larger velocity, amplitude, and accuracy than vertical ones. 

\paragraph{Movements of the left and right eyes} \label{movement} There are also separate groups of motor neurons involved in movements of the left and right eyes \citep{king2011binocular}. Eye dominance is referred to as laterality; approximately 70\% of people are right-eye dominant. 
The dominant eye 
executes eye movements faster and more accurately and leads the visual input \citep{tagu2016eye}. 

Natural attention mechanisms are often modeled in eye-tracking research to explain how individuals allocate their attentional focus during different tasks, 
especially decision-making tasks, which have been widely studied in psychological and behavioral research \citep{wedel2023modeling}.

\subsubsection{Attention Mechanisms in Deep Learning} \label{attentionDL}

Attention mechanisms are fundamental components in modern deep learning architectures. 
They allow models to dynamically weigh input sequences based on relevance, enabling nuanced representations of both intra- and inter-sequence relationships. In this work, we focus on three types of attention mechanisms: self-attention, cross-attention, and co-attention.

Self-attention captures dependencies within a single sequence by replacing a focal element in the sequence by a weighted average of the entire sequence, the weights being determined by the relevance of each token element in the sequence to the focal element \citep{vaswani2017attention}. It serves as the foundation for Transformer-based models, such as Chronos \cite{ansari2024chronos}, and is useful for learning contextual representations. Cross-attention computes attention across two (or more) sequences, where one sequence (e.g., queries) attends to another (e.g., keys and values). This asymmetric structure allows the model to integrate information from a distinct but related input. Co-attention mechanisms, in contrast, operate symmetrically on two (or more) sequences by applying parallel self-attention on each and modeling their coordinated behavior. Variants such as crossed co-attention allow each sequence to condition on the other in a bi-directional fashion.

In STARE, we leverage both cross-attention and co-attention to model spatio-temporal patterns in eye-tracking data (in addition to the self-attention mechanism in the Chronos/T5 backbone). Cross-attention enables the model to capture directional or asymmetric relationships. For instance, in binocular eye movement data, it captures interocular influences such as the dominant eye guiding fixation sequences. 
In fixation data, it models dependencies where horizontal (x-axis) movements often lead and influence vertical (y-axis) ones.

Co-attention is effective for modeling symmetric and coordinated behaviors. In binocular data, it captures the joint but decoupled control of both eyes, while in fixation data it reflects the coordination between horizontal and vertical movements, which are independently controlled yet interdependent.

Table~\ref{tab:cross_co_attention} summarizes the differences and similarities between these two attention mechanisms and illustrates the distinct roles that cross-attention and co-attention play in STARE when modeling eye-tracking data across different input modalities. 
These attention mechanisms are crucial in accurately modeling the spatio-temporal nature of eye movements.
The STARE model combines both cross- and co-attention to better understand gaze behavior and its influence on decision-making processes.

\begin{table*}
    \centering
    \renewcommand{\arraystretch}{1.3}
    \begin{tabular}{|l|p{7cm}|p{7cm}|}
        \hline
        \textbf{Feature} & \textbf{Cross-Attention} & \textbf{Co-Attention} \\
        \hline
        \textbf{Mechanism} & 
        Horizontal ($y^1$) eye movements are dominant and provide the context for vertical 
        ($y^2$) movements & 
        Horizontal ($y^1$) and vertical ($y^2$) eye movements are independently controlled yet coordinated\\
        \hline
        \textbf{Intuition} & When making vertical eye movements people pay attention to the horizontal positions in the scan path & When making horizontal (vertical) eye movements people pay attention to horizontal (vertical) positions in the scan path \\
        
        \hline
        \textbf{Input} & $y^1$-ROIs as \textbf{Query}, $y^2$-ROIs as \textbf{Key, Value} & Both $y^1$- and $y^2$-ROIs are \textbf{Query, Key, Value} \\
        \hline
        \textbf{Formulation} & $A = F \left (Q_{y^1}  K_{y^2}^T \right)\cdot V_{y^2}$ & $A_{y^1} = F\left (Q_{y^1}  K_{y^1}^T \right)\cdot V_{y^1}$ and $A_y=F\left (Q_{y^2}  K_{y^2}^T\right)\cdot V_{y^2}$ \\
       \hline
        \textbf{Direction} & One-way $( y^1 \rightarrow y^2)$ or $(y^2 \rightarrow y^1)$ & Mutual $(y^1 \leftrightarrow y^2 )$ \\
        \hline
        \textbf{When to Use?} & When \textbf{one coordinate} has a dominant influence on the other in the decision & When \textbf{both coordinates} are separately controlled but influence each other in the decision \\
        \hline
        \textbf{Output} & $y^1$-ROIs updated with $y^2$-context & $y^1$ \& $y^2$-ROIs refined together \\
        \hline

    \end{tabular}
    \caption{Comparison of Cross-Attention and Co-Attention on $y^1$ and $y^2$ inputs in STARE. $y = (y^1, y^2)$ represents x- and y-coordinates of fixations; for binocular gaze data with four channels, $y = (y^1,\dots,y^4)$, with similar attention mechanisms applied.}
    \label{tab:cross_co_attention}
\end{table*}

\subsection{LLMs for time series} 
Recent reviews of foundational, pre-trained, multivariate, and spatio-temporal deep learning models for time series data were provided by \cite{liang2024foundation}, \cite{ma2024survey}, \cite{shao2024exploring} and \cite{wang2020deep}. Models for time-series forecasting have involved Recurrent Neural Networks \cite{rangapuram2018deep, wang2019deep, salinas2020deepar}, Convolutional Neural Networks \cite{dong2024simmtm, wu2022timesnet, zhang2022self}, Diffusion Models \cite{wang2023diffload, wen2023diffstg} and Transformers \cite{lim2021temporal, nie2022time}. Here we focus on the latter. The Transformer \citep{vaswani2017attention} 
is an encoder-decoder LLM where the input sequence of tokens is mapped to a lower-dimensional embedding. 
The key innovation in the Transformer is the (multi-head) self-attention mechanism that enables it to capture dependencies between the input tokens and makes it a powerful tool for other tasks, including the forecasting time-series data. More details on the various attention mechanisms are provided in section \ref{Attention}. The work by \cite{unger2024predicting} uses a Transformer-based  
architecture to model time-series data of raw eye-movements for choice prediction, while \citep{wang2024gaze} infuse the Transformer-based BERT with fixation metrics to enhance Natural Language Processing. Other recent research has similarly applied pre-trained or fine-tuned Transformer-based LLMs to time-series data \citep{rasul2023lag,das2023decoder,woo2024unified,zhou2023one, jin2023time}. 

\subsection{Chronos} \label{subsec: chronos}
Whereas these prior time-series models predict a finite set of tokens, time series take on continuous values from an unbounded domain. Chronos \cite{ansari2024chronos} recognizes this feature and tokenizes the real values of univariate time series into discrete bins 
to allow the training of extant Transformer-based LLMs directly on these tokens without any modifications. Tokenization is an effective model adaptation strategy to represent time-series data as embeddings that are accessible to foundational models 
\cite{liang2024foundation}, such as the T5 Text-to-Text Transformer  \citep{raffel2020exploring} that is part of the Chronos architecture. Specifically, Chronos' tokenization involves (a) scaling each observation in the time series by the mean of the absolute value of all observations, (b) binning the scaled values by selecting bin centers uniformly on the domain of the data, (c) adding tokens to pad time series of different lengths to the same fixed length and to mark the end of the time-series. 

While Chronos offers a general foundational model for time-series data through lightweight tokenization and compatibility with pre-trained LLMs, other Transformer-based time-series models adopt more specialized strategies. The IBTSF model \cite{bhogade2024time} leverages a Transformer-based architecture to capture long-term dependencies and generate probabilistic forecasts; GRAFormer \cite{yang2024graformer} improves efficiency in multivariate time series forecasting by refining attention and feedforward components; and GridTST \cite{cheng2024leveraging} models multivariate time series using a structured grid of variate and time tokens. These methods perform well in their respective domains but are tightly coupled to specific architectures or assumptions that limit their applicability to non-standard inputs. We chose Chronos as a foundation because of its modular design, adaptability, and pre-training on diverse time-series datasets that enable its extension to new modalities and integration with large language models.

Although Chronos presents a powerful foundational model, it is not expected to capture eye-tracking time series well, even after fine-tuning, because (1) it is tailored to univariate time series while eye-tracking data has a spatio-temporal structure, (2) the tokenization of time series fails to adequately represent the multivariate, discrete-continuous nature of fixation-based eye-tracking data. Fixations are characterized by continuous-valued spatial coordinates and variable durations, forming irregular and fine-grained spatio-temporal sequences. Chronos’ tokenization strategy, designed for quantized trend patterns in univariate time series, aggregates over fixed intervals, and lacks the spatial granularity and temporal flexibility needed to preserve fixation-level resolution. This mismatch can lead to degraded representations that overlook essential gaze dynamics and patterns of visual attention. 

\subsection{Predicting choice using eye movements}
\label{subsec:predict-choice}
While most of the prediction models used in the prior eye-tracking literature are extensions of regression and logistic regression models, several studies use more advanced statistical models to predict human decision making from scan-paths. A Hidden Markov Model of scan-paths on ROIs was used by \citep{shi2013information} in the context of choices of products on a comparison website. In a similar setting, \cite{martinovici2023attention} use a regression spline model of gaze on product ROIs to forecast consumers' product choices. Reviews of these and other approaches using eye-tracking data to understand and predict choice decisions include \cite{orquin2013attention, wedel2023modeling}. 

Eye tracking data has also been used as input to NN models to predict users’ decisions. 
For example, decisions made by players in games were modeled using a single layer MLP \citep{krol2017}, 
CNNs were used to study strategic choices in economic games \citep{polonio2015strategic} and game players' decisions \cite{byrne2023predicting}, and LSTMs were used to model product choices in virtual environments \cite{palacios2023predicting}. This research often involves binary data or data on choices of one out of a relatively small number of choice options (4 or 5).  

A few studies involve the use of more advanced deep learning methods to predict decisions from eye movements. 
Most models used to date are domain-specific. To understand players' strategies in economic games, researchers in \cite{byrne2023predicting} transform raw eye-tracking data into images by overlaying ROI-based scan-paths on images of the decision environment, and use those as an input to a pretrained VGG-19 \citep{simonyan2014very} architecture to predict players' decision strategies. The transformer-based architecture RETINA, developed by \cite{unger2024predicting},  uses eye-movement data to predict consumers choice of one out of four laptops on a comparison website. Both studies demonstrate that more advanced deep learning architectures, such as CNNs and Transformers, have the ability to outperform established statistical and machine learning methods, 
bidirectional LSTM, 
and even BERT 
\citep{devlin2018bert, wang2024gaze}. 
Similarly, \cite{sims2020neural} use raw data 
to predict if a person working on a certain task is confused or not. 
To make such predictions, \cite{sims2020neural} combines GRU and CNN outputs and feeds them into a fully connected layer for binary confusion classification.
A similar architecture 
was used in \cite{sriram2023classification}, but for a different classification problem where it was finetuned to diagnose the Alzheimer disease. Note that the solutions reported in \citep{byrne2023predicting}, \citep{unger2024predicting}, and \cite{sims2020neural, sriram2023classification} overcome the challenge of limited data in eye-tracking studies by using raw eye movement data, rather than more widely used scan-paths of fixations. 

\begin{figure}
  \centering
  \includegraphics[width=1.0\linewidth]
  {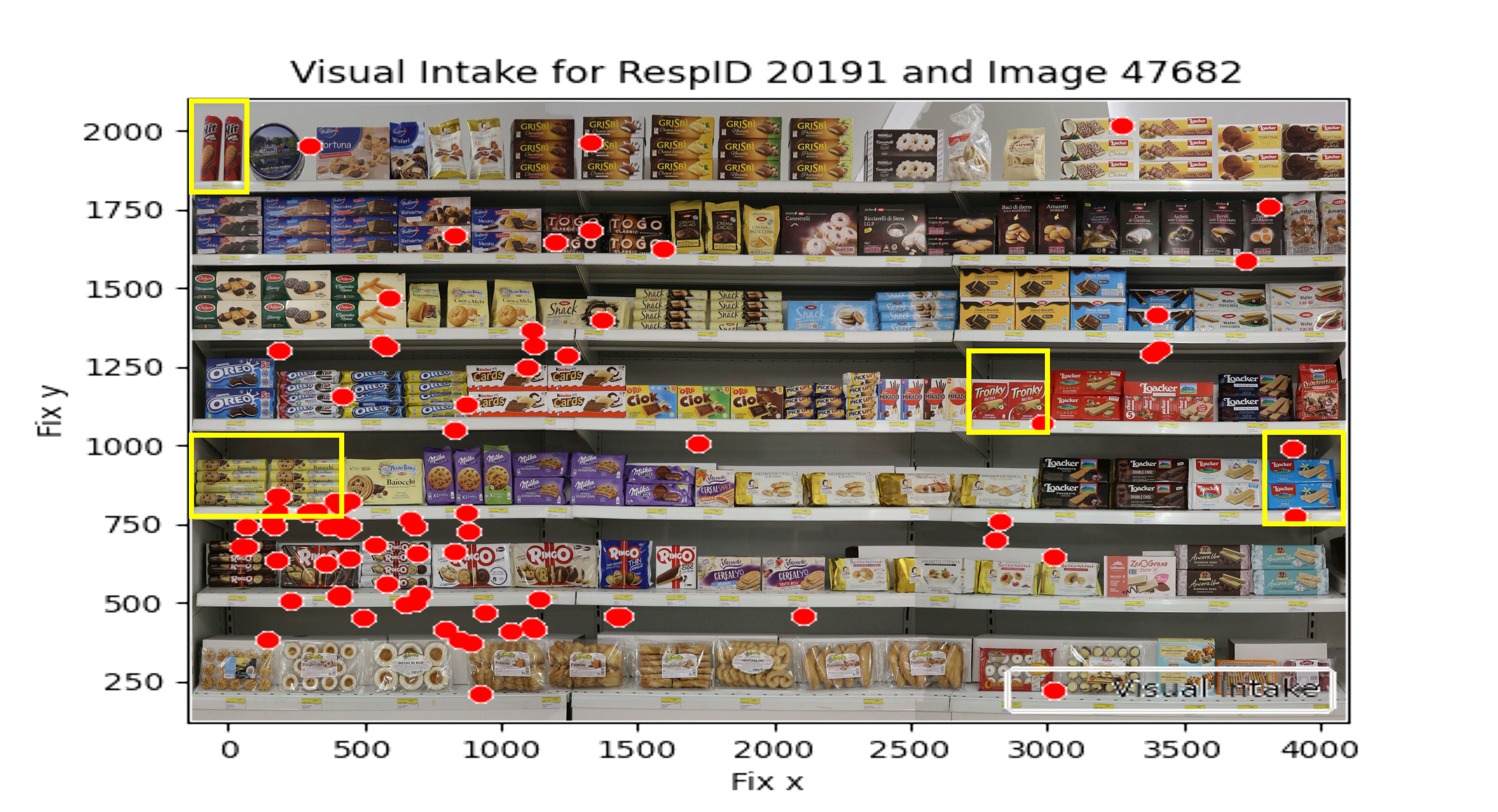}
  \caption{Sample image of a decision environment in visual intake (fixations) of one participant.}
  \label{fig:Fixations}
\end{figure}

\section{Method}

\subsection{Problem formulation}
Scan paths of the eyes during people's decision making provide information on the unobserved decision process and are predictive of decision outcomes \citep{wedel2023modeling}. However, good prediction requires an accurate extraction of the right signals from the right sources, which is more difficult due to the 
noisy \emph{bivariate, spatio-temporal} eye-tracking data, for which few AI-based methods, including foundational LLM models, have demonstrated satisfactory performance \citep{unger2024predicting, wang2024gaze}. 

In contrast, the Chronos framework presents a DL model developed for univariate time series data that deploys a time series-based tokenization method and the self-attention mechanism to capture temporal dependencies in time series. 
However, although Chronos offers potential to model scan-paths, its tokenization \emph{is not well-suited for spatio-temporal eye fixation data} because of the multivariate discrete-continuous nature of the time series of eye-movement data, as explained in Section~\ref{overview-eyetrackers-eye-movements}. In addition, the standard univariate self-attention mechanism is insufficient to capture the dependencies in bivariate spatio-temporal eye-movement data. 

To address these limitations, the goal of the present work is, first, to develop a new tokenization strategy 
for the prediction of decisions from eye-movement data. To do this, we replace Chronos' tokenization with one based on the Regions of Interest (ROIs) on images of the decision environment that capture human visual attention during decision making. 
Second, we significantly extend the Chronos framework by including, on top of the token embeddings produced by Chronos' T5 component, the DL-based attention mechanisms for bivariate time series inspired by various human neuro-motor and attention processes involved in eye movements, as explained in Section~\ref{overview-eyetrackers-eye-movements}. We describe our model, including our tokenization method and a broad overview of the STARE method, in the next subsection. We then describe the specifics of the directional attention mechanisms, as applicable to eye-tracking, in Section~\ref{Attention} and the end-to-end training procedure of STARE (Algorithm 1) in Section~\ref{subsec: end-to-end}. Finally, we compare the performance of STARE relative to several alternative approaches in Section~\ref{section:experiments}. 

\subsection{Method framework}
\label{Method-framework}

\begin{figure}[t]
\centering
  \includegraphics[width=1.0\linewidth]
  {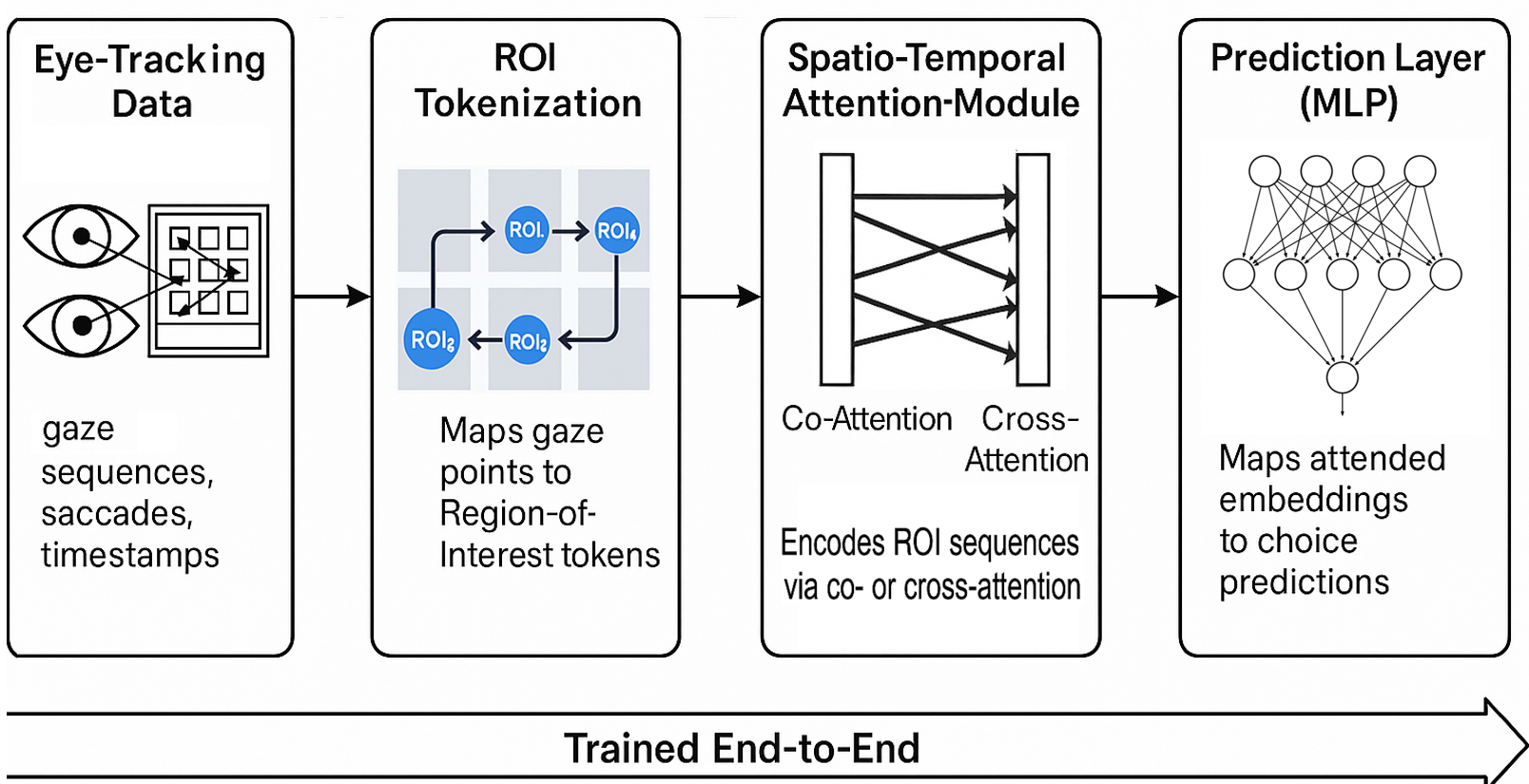}
  \caption{STARE framework: End-to-end pipeline for transforming raw eye-tracking data into decision-aware predictions using ROI tokenization, spatio-temporal attention, and MLP classification.}
  \label{fig:stare_framework}
\end{figure}

Our approach builds on the Chronos \cite{ansari2024chronos} architecture via a novel tokenization strategy tailored for the eye-tracking domain.
Unlike Chronos and other foundational time-series models \cite{liang2024foundation} that generate tokens based on statistical properties of sequential data, our method tokenizes \emph{individual-level spatio-temporal scan paths} by identifying semantically meaningful \emph{ROIs} \cite{orquin2016areas} within the decision environment (e.g., an image of a supermarket shelf) as described below. The main symbols used in this paper are summarized in Table~\ref{tab:symbol_tab}.  

As illustrated in Figure~\ref{fig:stare_framework}, our proposed method STARE follows a structured, end-to-end pipeline for transforming raw eye-tracking data into decision-aware representations for user choice prediction. The pipeline begins by converting raw fixation streams into ordered ROI tokens, which, after being passed through Chronos' pre-trained T5 encoder to generate contextualized embeddings, are then processed through attention-based DL mechanisms designed to capture spatial and temporal dependencies in gaze behavior. These attention-informed embeddings are finally mapped to user choices through a multi-layer perceptron (MLP) prediction head. The entire architecture is trained end-to-end to jointly optimize gaze encoding and decision inference. We now describe each component of the framework in detail, with the ROI definition in Section~a, its tokenization in Section~b, and the attention mechanisms in Section~\ref{Attention}. Finally, the pseudo-code of the STARE algorithm is outlined in Section~\ref{subsec: end-to-end}.

\paragraph{Regions of Interest (ROI)} \label{ROIs}
\emph{ROIs} 
can be defined in various ways, for example by annotating objects and other semantic structures, or via a spatial grid imposed upon images of the decision environment. Thus, researchers may annotate regions of an image 
containing semantically meaningful information that may attract visual attention, rather than deal with abstract collections of pixels. For example, in our data on online selling of Dell laptops, ROIs on a row $\times$ column (attributes $\times$ products) display, consist of rectangular regions on the Dell webpage either identifying an image of a Dell laptop or semantically meaningful textual descriptions of laptops' various attributes, such as CPU, memory, hard disk, etc. \cite{shi2013information}. 
In the setting of our other dataset, where the goal is to predict which products users will choose from a dense visual display, representing shelves in a supermarket, ROIs may correspond to individual products placed on these shelves, branded sections, or broader shelf zones. For example, some ROIs are shown as yellow rectangles in Figure~\ref{fig:Fixations}.

\paragraph{Tokenization} \label{tokens} The existing Chronos tokenization process was developed for univariate time series data,
as described in Section~\ref{subsec: chronos}. 
Although it could be applied to bivariate (x-y) time series of fixations by scaling and quantization of the inputs for the x- and y- input channels independently,
this approach does not heed spatial semantic relationships in the decision environment, nor the discrete/continuous nature of the eye-movement data. 
Therefore, for the \emph{STARE} architecture we propose a novel tokenization strategy for spatio-temporal fixation data, which involves mapping the \emph{continuous} x- and y- coordinates of eye fixations onto the indices of ROIs defined on the image of the visual stimulus.
Figure~\ref{fig:stare_framework} illustrates the STARE tokenization strategy, where the x- and y- dimensions of the time series of fixations are mapped to ROIs on the image of the decision environment, which leads to the sequences of ROI tokens. Alternatively, when using raw gaze data to analyze the scan-paths of both eyes separately, the time series of the left and right eyes are each mapped to ROIs.

Specifically, first, ROIs are ``manually" annotated on images of the decision environment and indexed, as shown in Figure \ref{fig:Fixations} (see yellow boxes in Figure 2). Second, the bivariate spatio-temporal input time series is scaled and transformed into a semantically meaningful time series of ROIs by mapping the x- and y- coordinates of the fixations of each person separately to their corresponding ROI-indices. 
Finally, the \emph{sequences of ROI tokens} are obtained as follows. Let $\left(y_{i,t}\right)_{t=1}^{T_i}$ be the bivariate time series of pixel-level fixations for $T_i$ fixations made by person $i$, and $R\left(y_{i,t}\right)$ be the ROIs containing pixel coordinates $y_{i,t}$, then the bivariate time series of pixel coordinates is mapped on to the bivariate sequence of tokens $R\left(y_{i,t}\right)_{t=1}^{T_i}$ on which Chronos' pre-trained T5 LLMs operate by encoding each token sequence, as described below and in Algorithm 1.

Note that this tokenization process may require a step in which a set of mutually exclusive and collectively exhaustive ROIs, covering the whole image, are predefined. Defining ROIs is often required to ground fixation data in object-level semantics, which in turn enables training of meaningful prediction models. For example, similar to approaches such as FixTag \cite{munn2009fixtag}, ROI information can be provided via a structured file that includes semantic labels and spatial coordinates, facilitating downstream learning tasks. Moreover, this process is scalable: in domains like retail shopping, visual layouts of shelves are repetitive, allowing annotations to be reused across many instances. 
Moreover, the ROI generation process can be partially or fully automated using standard computer vision methods, such as object detection or segmentation based on product metadata \cite{gundimeda2018automated}, making the whole annotation process tractable and scalable in practice. All this makes the ROI definition step a practical and domain-aware mechanism to inject semantic structure into the tokenization process.

The advantage of the STARE tokenization strategy is that, while Chronos produces tokens based on the numerical properties of the time series, STARE tokens are based on a mapping of fixation coordinates to ROIs that are semantically meaningful relative to the visual stimulus and the participants' attention processes during decision making. Although the ROIs can be defined in various ways, as was explained earlier, 
in our setting, where the goal is to predict which products users will choose from a dense visual display, such as shelves in a supermarket, ROIs may correspond to individual products, branded sections, or broader shelf zones. This allows \emph{tokenization to reflect semantically meaningful objects} rather than abstract pixel coordinates.

In summary, the proposed tokenization allows us to effectively compress spatio-temporal eye-movement time-series data into compact, decision-relevant representations that align with meaningful business and behavioral constructs. This enables 
improved prediction of choice decisions, as is shown in Section~\ref{results}.

\begin{table}[t]
    \scriptsize
    \renewcommand{\arraystretch}{1.4}
    \setlength{\tabcolsep}{6pt}
    \centering
    \begin{tabularx}{\columnwidth}{|>{\raggedright\arraybackslash}m{2.4cm}|>{\raggedright\arraybackslash}X|}
    \hline
    \textbf{Symbol} & \textbf{Description} \\
    \hline
    $m \in \{\text{class}, \text{count}\}$ & Task type: binary choice (class) or choice frequency (count) \\
    \hline
    $i=1,\cdots, N$ & Indexes respondents \\
    \hline
    $j=1,\cdots, J$ & Indexes the choice alternatives; $J$ is the number of products shown \\
    \hline
    $t=1,\cdots, T_i$ & Time index: equally spaced gaze points or fixations for person $i$ \\
    \hline
    $k=1,\cdots,K$ & Input channels: x-y fixations or binocular gaze coordinates \\
    \hline
    $d$ & Embedding dimensionality \\
    \hline
    $u_i, \hat{u}_i$ & True and predicted outcome vector over $J$ items \\
    \hline
    $y^k_{i,t}$ & $k$-th input channel at time $t$ \\
    \hline
    $y_{i,t}=\left(y^{k}_{i,t}\right)_{k=1}^K$ & $K$-dimensional gaze input for subject $i$ \\
    \hline
    $\mathcal{G}_i := \left(y_{i,t}\right)_{t=1}^{T_i}$ & Gaze time series for subject $i$ \\
    \hline
    $r^k_t = \mathcal{R}\left(y_{i,t}\right)$ & ROI token for pixel $y_{i,t}$ in channel $k$ \\
    \hline
    $\mathcal{T}^k=\left(r^k_t\right)_{t=1}^{T_i}$ & ROI token sequence for input channel $k$ \\ 
    \hline
    $\mathcal{D} = \left(\mathcal{G}_i, u_i\right)_{i=1}^{N}$ & Training data: gaze input and outcomes \\
    \hline
    $H^k=\left(h^k_t\right)_{t=1}^{T}$, $h^k_t \in \mathbb{R}^{d}$ & T5 embedded ROI token sequence for channel $k$ \\
    \hline
    $\mathbf{Z}^{\text{co}}=\left( \mathbf{Z}^{\text{co},k}\right)_{k=1}^K$ & Co-attention aware embeddings of input channels \\
    \hline
    $\mathbf{Z}^{\text{cross}}, \mathbf{Z}$ & Cross-attention and combined embeddings \\
    \hline
    $\mathcal{L}, \nabla_\theta \mathcal{L}, \eta, \theta $ & Loss function, gradient, learning rate, and model parameters \\
    \hline
    \end{tabularx}
    \caption{Symbol Table}
    \label{tab:symbol_tab}
\end{table}
\vspace{-1.5ex}

\subsection{Directional Attention for Dominant Fixations}
\label{Attention}

As explained in Section~\ref{overview-eyetrackers-eye-movements}, the physiology and neurobiology of the eyes and their movements teach us that the right and left eyes move somewhat differently for most people. Therefore, to make better choice predictions, in addition to tracking the exact x- and y-locations of eye fixations, it is useful to track the scan paths of \emph{both} eyes. However, most eye movement analyses calculate fixation points as the average of the fixation locations of the left and right eye. Based on the material presented in Section~\ref{overview-eyetrackers-eye-movements}, we maintain that this averaging may overlook important ocular dominance cues that are relevant to decision making. Therefore, in this paper, we propose to \emph{explicitly model eye movements in both horizontal (x) and vertical (y) directions, as well as from both the left and right eyes whenever possible, using co- and cross-attention mechanisms of Deep Learning}, as is explained further in this section. This design choice is motivated by the fact that ocular or directional dominance affect how users visually explore complex stimuli, such as shelves filled with products. The co- and cross-attention mechanisms of Deep Learning \emph{are well-suited} to model these human neuro-motor and attention mechanisms of eye movements, as is explained in this section. We test these mechanisms on two datasets that comprise fixations and also raw gaze points of both eyes.

As summarized in Table~\ref{tab:cross_co_attention}, co-attention treats two sequences of ROIs both as Query, Key, and Value, allowing them to refine each other symmetrically. The model applies such mutual attention, ensuring that \emph{both inputs contribute equally} without one dominating the other, to represent eye movement behavior. The attended outputs are then combined and passed to the classification model.

As an example, we could explore the co-attention mechanism in tracking the dependencies between the x- and y-coordinates of the eye focus, which allows the model to capture spatial dependencies in gaze patterns corresponding to horizontal and vertical eye movements, since x- and y-sequences might influence each other equally, with no predefined dominance. Similarly, we can explore the co-attention mechanism for the left- and right eye, which implies that the left- and right eye movement sequences may influence each other equally, with no particular dominance.

As discussed in Section~\ref{overview-eyetrackers-eye-movements}, the cross-attention mechanism assumes that one input influences the other when computing attention, implying the directional influence of one input on another \cite{vaswani2017attention}. Unlike co-attention, cross-attention ensures that one input refines another \textit{with a predefined dominance}. Cross-attention differs from co-attention in its directionality: one sequence influences the other in a one-way manner.

As an example, an eye movement domain well suited for exploring the cross-attention mechanism constitutes the directional influence that one eye may exert over the other during decision-making; 70\% of people are right-eye dominant, 
which makes them execute right-eye movements faster and more accurately, 
as explained in Section~\ref{movement}. In this cross-attention setup, which models the directional influence that one eye may exert over the other, one eye’s ROI sequence is treated as the Query (Q), representing the primary source of information, while the other eye’s sequence serves as the Key (K) and Value (V), thus providing contextual grounding. Specifically, we tested two configurations: one where the right-eye sequence acts as the query and the left-eye sequence as key and value (representing right-eye dominance), and another where the roles are reversed (representing left-eye dominance). This allows us to examine whether directional asymmetries, such as those associated with ocular dominance, have a measurable impact on decision prediction performance. 


When analyzing \textit{fixation data} (i.e., without binocular separation), we implement the cross-attention mechanism by treating the x-axis sequence as the Query and mapping it to all vertically aligned ROIs (i.e., the full column of a shelf) that correspond to the same horizontal position. Similarly, the y-axis sequence serves as the Key and Value, referencing all horizontally aligned ROIs (i.e., a row of the shelf) at the corresponding vertical position. This setup enables the model to capture directional interactions between gaze shifts across rows and columns, thus reflecting how users visually explore structured decision environments, such as product shelves.

For both fixation data and raw gaze data, we simultaneously implement the co-attention and cross-attention mechanisms to operate simultaneously, allowing for coordinated and/or dominant influences of directionality or laterality in eye movements. 



\subsection{End-to-End Training} \label{subsec: end-to-end}





\begin{algorithm}
\caption{\textsc{STARE} Training Procedure (Fixation Data)}
\label{alg:stare-train}
\begin{algorithmic}[1]
\State \textbf{Input:} Training set $\mathcal{D} = \left(\mathcal{G}_i, u_i\right)_{i=1}^{N}$, where $\mathcal{G}_i := \left(y_{i,t}\right)_{t=1}^{T_i}$ is a bivariate (K=2) fixation time series; ROI map $\mathcal{R}$, task type $m \in \{\text{class}, \text{count}\}$; learning rate $\eta$
\State \textbf{Output:} Trained model parameters $\theta$

\For{each epoch}
    \For{each $(\mathcal{G}_i, u_i) \in \mathcal{D}$}
        \State \textbf{Tokenization:}
        \For{each fixation $y_{i,t} = (y^1_{i,t}, y^2_{i,t})$ in $\mathcal{G}_i$}
            \State $r^1_t \gets \textsc{MapXtoVerticalROIs}(y^1_{i,t}, \mathcal{R})$
            \State $r^2_t \gets \textsc{MapYtoHorizontalROIs}(y^2_{i,t}, \mathcal{R})$
        \EndFor
        \State $\mathcal{T}^1 \gets (r^1_t)_{t=1}^{T_i}$ \Comment{X-axis ROI token sequence}
        \State $\mathcal{T}^2 \gets (r^2_t)_{t=1}^{T_i}$ \Comment{Y-axis ROI token sequence}

        \State \textbf{T5 Embedding:}
        \State $\mathbf{H}^1 \gets \textsc{T5Embed}(\mathcal{T}^1)$
        \State $\mathbf{H}^2 \gets \textsc{T5Embed}(\mathcal{T}^2)$

        \State \textbf{Attention:}
        \State \textbf{Co-Attention:}
\State $\mathbf{Z}^{\text{co},1} \gets \textsc{Attention}(\mathbf{H}^1, \mathbf{H}^2, \mathbf{H}^2)$ 
\State $\mathbf{Z}^{\text{co},2} \gets \textsc{Attention}(\mathbf{H}^2, \mathbf{H}^1, \mathbf{H}^1)$ 
\State $\mathbf{Z}^{\text{co}} \gets \textsc{Combine}(\mathbf{Z}^{\text{co},1}, \mathbf{Z}^{\text{co},2})$

\State \textbf{Cross-Attention:}
\State $\mathbf{Z}^{\text{cross}} \gets \textsc{Attention}(\mathbf{H}^1, \mathbf{H}^2, \mathbf{H}^2)$ 

\State \textbf{Fusion:}

\State $\mathbf{Z} \gets \textsc{Combine}(\mathbf{Z}^{\text{co}}, \mathbf{Z}^{\text{cross}})$

        \State \textbf{Prediction:}
        \If {$m = \text{class}$}
            \State $\hat{u}_i \gets \textsc{Softmax}(\mathbf{Z})$
            \State $\mathcal{L} \gets \textsc{CrossEntropy}(\hat{u}_i, u_i)$
        \ElsIf {$t = \text{count}$}
            \State $\hat{u}_i \gets \textsc{Linear}(\mathbf{Z})$
            \State $\mathcal{L} \gets \textsc{MSE}(\hat{u}_i, u_i)$
        \EndIf

        \State \textbf{Update:} $\theta \gets \theta - \eta \nabla_\theta \mathcal{L}$
    \EndFor
\EndFor

\end{algorithmic}
\end{algorithm}

In this section we describe the
\emph{Structured Token-Aware Representation of Eye-tracking} (\textsc{STARE}) algorithm, both its training and inference components that are presented as Algorithm~\ref{alg:stare-train}, for fixation data (we describe its modifications for raw binocular gaze data below). We describe the algorithm for \textit{fixation data} first. Conceptually, at a high level, \textsc{STARE} works as follows. First, it transforms fixation-based time series of $x$- and $y$-coordinates of the eye-movement data into Region-of-Interest (ROI) token representations, as was described in Section~\ref{tokens}. These tokens are then embedded using the Chronos' pre-trained T5 encoder and passed through co-attention and/or cross-attention mechanisms to form the final embeddings used in the task of subsequent predictions or decision making. The specifics of this process are further described in Algorithm 1. 

The pipeline of Algorithm 1 consists of four main stages: Tokenization, Embedding, Attention, and Prediction, which are described below.
Let $\mathcal{G}_i := (y_{i,t})_{t=1}^{T_i}$ denote the bivariate time series of fixations for user $i$, where each $y_{i,t} = (y^1_{i,t}, y^2_{i,t})$ corresponds to the x- and y-coordinates of the pixel-level gaze position recorded at fixation $t$. Note that the fixations $t$ are unequally spaced and the sequence length $T_i$ varies across users depending on gaze behavior. Each training instance in the dataset is represented as a pair $(\mathcal{G}_i, u_i)$, where $u_i$ is the supervision target, a class label for classification tasks or a count for regression tasks.

Algorithm 1 begins with the tokenization step (row 5), where we extract for each user $i$ a sequence $\left(y_{i,t}\right)_{t=1}^{T_i}$ of bivariate time series of $T_i$ pixel-level fixations by collecting their gaze points $y_{i,t}$ and ordering them based on the time they occurred. This creates a time-based fixation sequence that reflects the user's visual exploration pattern, capturing how their eyes moved across the screen over time. These fixations are discretized using a \textit{ROI map} $\mathcal{R}$, which partitions the visual scene into a fixed grid of non-overlapping regions, as explained in Section~\ref{tokens}. 
The ROI map is exhaustive and mutually exclusive, ensuring that each fixation is mapped to exactly one ROI. 
More specifically, fixation coordinates are discretized using two routines: \textsc{MapXtoVerticalROIs} (row 7) and \textsc{MapYtoHorizontalROIs} (row 8), which map the x- and y-coordinate streams separately to vertical and horizontal ROI tokens based on their spatial alignment in $\mathcal{R}$. This produces two discrete token sequences, $\mathcal{T}^1$ and $\mathcal{T}^2$, which preserve the temporal dynamics of gaze behavior (rows 9 and 10).

After tokenization, each ROI token sequence is then passed through the pre-trained T5 encoder via \textsc{T5Embed} (rows 12–13) in the Embedding stage, which transforms the symbolic ROI tokens into contextualized embedding vectors. For a sequence of tokens $\mathcal{T} = (r_1, \ldots, r_T)$, the output of the \textsc{T5Embed} encoder is a sequence of hidden states $\mathbf{H} = (\mathbf{h}_1, \ldots, \mathbf{h}_T)$, where each $\mathbf{h}_t \in \mathbb{R}^d$ encodes both the semantic identity of the fixated region and its contextual relationship to surrounding regions.

To model interactions between the embedding sequences $\mathbf{H}^1$ and $\mathbf{H}^2$, \textsc{STARE} supports two attention mechanisms in the Attention stage. The co-attention mechanism captures mutual influences between the $x$- and $y$-coordinate embeddings by allowing each to attend to the other symmetrically (row 16-18). In contrast, the cross-attention mechanism emphasizes directional influence, as explained in Section~\ref{Attention}, where the $x$-embedding sequence attends to the $y$-embedding sequence (row 20). These complementary views of gaze interaction are fused using the \textsc{Combine} operation in row 22, implemented as concatenation, which aggregates the outputs of both mechanisms into a unified representation $\mathbf{Z}$. This joint attention strategy is designed to reflect scenarios in which spatial eye movements exhibit both coordinated and dominant directional patterns, thereby enabling richer modeling of the decision process.

STARE supports two types of prediction tasks in the Prediction stage. In the \textit{classification task}, such as predicting selected products on a shelf, $u_i$ is $J-$dimensional, and the model uses Softmax activation and is trained with the cross-entropy loss function (rows 25 - 26). In the \textit{regression task}, such as estimating the number of selected items, $u_i$ is 1-dimensional, and the model applies a linear output layer and is trained with mean squared error (MSE) loss (rows 28 - 29). The model is trained end-to-end via the stochastic gradient descent method, iterating across all samples to update parameters $\theta$ (row 30). 

The inference procedure 
uses the same forward pass logic as in Algorithms 1, however, without updating the model parameters (i.e., without rows 26, 29 and 30 in Algorithm 1). It predicts $\hat{u}$ without computing the loss function $\mathcal{L}$ (rows 26 and 29 of Algorithm 1).

In the case of \textit{raw binocular gaze data} the algorithm is similar, but each $y_{i,t} = \left(y^k_{i,t}\right)_{k=1}^4$ corresponds to the x- and y- coordinates of the gaze position of the left and the right eyes in pixel coordinates, respectively. Since the data comprises raw samples taken by the eye tracker at a frequency of 50Hz, the time points are equally spaced, but the sequence length $T_i$ varies across users due to variations in gaze behavior. 

\subsection{Training} \label{subsec: training}
We implement the STARE model based on Tensorflow using end-to-end training. To optimize the network parameters, we used the Adam optimizer with a maximum learning rate of 2e-5 and a decay factor of 0.5. For binary prediction tasks, we employ cross-entropy loss, while mean squared error (MSE) is used to predict the number of items chosen from a shelf. The dimension of the token embeddings is set to 768. To determine the optimal hyperparameters, we conduct a grid search. Additionally, we apply an early stopping strategy with a patience of 5 epochs to prevent overfitting and determine the appropriate number of training epochs. Unless early stopping intervened, training was conducted for up to 1000 epochs. For ROI token representation, we use the T5 Small model, selected for its strong zero-shot performance and relatively low training cost \cite{ansari2024chronos}, making it a practical choice for efficient and scalable training. 



\section{Experimental Settings}
\label{section:experiments}

In this section we describe the data and the experimental settings in subsection~\ref{subsection: data}, the comparison benchmarks in subsection~\ref{subsection: benchmarks} and the performance metrics used in the comparison analysis in subsection~\ref{subsection: protocol}. We draw upon two datasets, the first comprising of monocular eye fixation locations on digital retail shelves recorded while participants could pick any number of products from the shelf, the second comprising of raw binocular gaze points on a comparison website recorded during a choice of one out of four laptops. 

\subsection{Data and Experimental setting} 
\label{subsection: data}

\paragraph{PRS-in-Vivo Retail data} This data was collected by the market research company PRS in Vivo\footnote{https://www.prs-invivo-group.com/} on $N=163$ participants. The dataset includes $J=244$ unique products shown in 901 ``facings" (i.e., different package views) across 8 digital supermarket shelves. In total, 701 purchases are recorded in 308 sessions. Purchase decisions by 163 participants involve indicating the pick of any number of the 244 products (or making no purchase at all). While making purchase decisions, eye-fixation data were collected with Tobii eye trackers (fixation number, start and end timestamp of fixation, fixation duration, x- and y-fixation coordinates)\footnote{http://www.Tobii.com}. ROIs, capturing non-overlapping product facings on the shelf, were manually annotated on the images of each shelf. There are 112.6 ROIs on average per shelf and a total of 901 ROIs on 8 shelves. There are on average 26.87 fixations per participant, a total of 16,409 observations, and 8,277 fixations. Figure~\ref{fig:Fixations} shows a sample fixation pattern for one participant and one shelf.\footnote{This dataset is provided under NDA and cannot be made public.} 

Note that choice predictions for the PRS-in-Vivo dataset are very challenging because of the very large set of choice options from which participants may choose any number of options. Similar data arise in many real-life decision tasks, where people are not constrained in choosing only a single option, but may pick \textit{any number} out of the available choice options, as in our supermarket case. For example, people commonly download multiple songs from digital music catalogs, 
choose assortments consisting of multiple magazines, 
consume news from several media platforms, 
and own multiple financial products from a financial institution. 
In addition to predicting \emph{each individual's choice outcome(s)}, we also predict the \emph{total number of items} each respondent chooses from the shelves. 

\paragraph{Laptop data} This data was collected by \cite{shi2013information} on $N=112$ participants making choice decisions of selecting one among four desktop computers. The display contained 12 attributes verbally describing the four computers; see Figure 3 of \cite{unger2024predicting}. Tobii infrared eye-tracking equipment was used to record eye movements, while participants made their decision. The data contains the x- and y-coordinates of all gaze points of the left and right eyes of participants collected via the eye tracker at a frequency of 50Hz, and the final choice decision ($j=1,\cdots,4$). This dataset thus contains raw eye-movement data (not the fixation data as in the PRS-in-VIVO dataset). In total, 52 non-overlapping contiguous ROIs were annotated manually on the images of the decision environment, capturing each attribute for all four products. Summary of the data is provided in \cite{shi2013information}.

The two datasets allow us to test and compare our proposed STARE model under different conditions. The PRS-in-Vivo dataset has two input channels, the x- and y-coordinates of the eye fixations in the scan-path for each participant. The fixations are unequally spaced in time. The laptop data has four input channels, the x- and y-coordinates of the raw gaze points of the left and right eyes for each participant, which are equally spaced (at intervals of 20 msec.).

\subsection{Baseline Comparisons} 
\label{subsection: benchmarks}
We compare our STARE approach with five state-of-the-art deep learning baseline models Bi-LSTM, BERT, Gaze-Infused BERT, RETINA, and Chronos, of which the last four are based on the Transformer approach. Collectively, the selected baselines described below provide a wide sample of SOTA alternatives:

\begin{itemize}
    \item \textbf{Bi-LSTM}: a recurrent neural network that processes the input sequence in both forward and backward directions. We use a one-layer bidirectional LSTM (Long-Short-Term-Memory) architecture \citep{hochreiter1997long}. 
    \item \textbf{BERT}: An extension of the BERT \citep{devlin2018bert} architecture that integrates human gaze signals into the self-attention mechanism. Each (x,y) gaze point is treated as a token in the sequence, mapped into an embedding vector via linear projection, and combined with positional embeddings to preserve temporal order. The sequence is processed by transformer layers, and the pooled output is passed to a classification head to predict the final choice outcome.
    \item \textbf{Gaze-Infused BERT \citep{wang2024gaze}}: A model that integrates human gaze information into BERT to enhance Natural Language Processing (NLP) performance. It constructs a linear weighted gaze representation by combining five fixation-based gaze signals: FFD (First Fixation Duration), GD (Gaze Duration), GPT (Go-Past Time), TRT (Total Reading Time), and nFix (Number of Fixations). The entropy weight method is applied to determine the contribution of each signal, and the resulting representation is embedded into the transformer encoder to enrich contextual understanding during self-attention. Since these gaze signals are computed from fixation data, they can't be used for raw gaze data. We use Gaze-Infused BERT only for the fixation-based experiments, where the final gaze embedding serves as the input for both choice and item count prediction tasks.
    \item \textbf{RETINA}, the architecture developed by \cite{unger2024predicting} and reviewed in Section~\ref{subsec:predict-choice}, which involves a combination of a Transformer architecture with a self-attention mechanism and metric learning to capture left-and right eye data streams to predict choice decisions from raw eye-tracking data. It is applied here to the single-channel fixation time series input and to the raw eye-tracking data of both eyes. 
    \item \textbf{Chronos}, the T5-based architecture developed by \cite{ansari2024chronos}, with tokenization via scaling and binning of the x- and y- input channels separately, as reviewed in Section~\ref{subsec: chronos}. 
\end{itemize}

\subsection{Performance Metrics and Comparison Protocol}
\label{subsection: protocol}
Evaluations were done using the 10-fold Cross-Validation with 30\% - 70\% splitting (30\% test and 70\% training data). Random negative sampling was used for binary predictions (1- buy, 0-not buy) for each shelf. 
We used three performance metrics for binary predictions, i.e., the accuracy, F1, and the AUC-ROC metrics. 
To assess the accuracy of the number of items chosen on each shelf, we employed three standard metrics: Root Mean Squared Error (RMSE), Mean Absolute Error (MAE), and Mean Absolute Percentage Error (MAPE\%). While RMSE and MAE quantify absolute deviations from the true values, MAPE expresses the average error as a percentage, facilitating interpretation across different scales.

\begin{table*}[t]
\centering
\scriptsize
\renewcommand{\arraystretch}{1.3}
\setlength{\tabcolsep}{5pt}
\begin{tabular}{|l|ccc|ccc|ccc|}
    \hline
    & \multicolumn{3}{c|}{\textbf{Choice Prediction (Fixation)}} 
    & \multicolumn{3}{c|}{\textbf{Item Count Prediction (Fixation)}} 
    & \multicolumn{3}{c|}{\textbf{Choice Prediction (Raw Gaze)}} \\
    \cline{2-10}
    \textbf{Model} 
    & \textbf{Accuracy} & \textbf{AUC-ROC} & \textbf{F1} 
    & \textbf{RMSE} & \textbf{MAE} & \textbf{MAPE (\%)} 
    & \textbf{Accuracy} & \textbf{AUC-ROC} & \textbf{F1} \\
    \hline
    Bi-LSTM 
    & 0.563 & 0.556 & 0.548 
    & 0.881 & 0.879 & 28.47 
    & 0.772 & 0.741 & 0.767 \\
    
    BERT 
    & 0.569 & 0.560 & 0.555 
    & 0.875 & 0.872 & 28.25 
    & 0.781 & 0.755 & 0.771 \\

    Gaze-Infused BERT 
    & 0.571 & 0.563 & 0.559 
    & 0.870 & 0.866 & 28.05 
    & -- & -- & -- \\
    
    RETINA 
    & 0.574 & 0.571 & 0.568 
    & 0.826 & 0.825 & 26.72 
    & \underline{0.789} & \underline{0.764} & \underline{0.778} \\
    
    Chronos 
    & \underline{0.597} & \underline{0.591} & \underline{0.589} 
    & \underline{0.817} & \underline{0.819} & \underline{26.53} 
    & 0.783 & 0.758 & 0.772 \\
    
    \textbf{STARE} 
    & \textbf{0.628} & \textbf{0.616} & \textbf{0.615} 
    & \textbf{0.761} & \textbf{0.756} & \textbf{24.49} 
    & \textbf{0.798} & \textbf{0.778} & \textbf{0.790} \\
    
    \hline
    \textbf{$\Delta$ vs. Best Baseline (\%)} 
    & \textbf{5.19\%} & \textbf{4.23\%} & \textbf{4.41\%} 
    & \textbf{6.86\%} & \textbf{7.70\%} & \textbf{7.68\%} 
    & \textbf{1.14\%} & \textbf{1.83\%} & \textbf{1.54\%} \\
    \hline
\end{tabular}
\caption{Performance comparison of benchmark models across fixation-based PRS-in-Vivo data and raw gaze laptop data. Fixation-based evaluations include both choice prediction and item count prediction, while raw gaze evaluations include only choice prediction. Best results are bolded; second-best results are underlined.}
\label{tab:benchmark_stare}
\end{table*}


%
\section{Results} 
\label{results}

We first compare the performance of STARE with benchmark models on the two datasets described in Section~\ref{section:experiments}, evaluating performance on both choice prediction and item count prediction tasks. We then present a series of ablation studies to assess the contributions of key components of our framework to its overall prediction accuracy. Finally, we analyze how the model’s performance varies with the viewing time, i.e., the duration the subject spends examining an image before making a selection.

\subsection{Benchmark Comparison on Choice and Item Count Prediction}

To evaluate the effectiveness of our \textsc{STARE} method, we conducted experiments on the PRS-In-Vivo dataset across two complementary prediction tasks: (1) binary choice prediction, i.e., whether a product is chosen, and (2) item count prediction, i.e., how many products are selected from a shelf during a shopping trip. In the PRS-In-Vivo experiment, participants were free to choose any number of items from each shelf, creating natural variation in both choice and frequency of selection (see Section~\ref{subsection: data}). Table~\ref{tab:benchmark_stare} reports the performance of \textsc{STARE} alongside the baseline models, including Bi-LSTM, BERT, Gaze-Infused BERT, RETINA, and Chronos, evaluated across both tasks.

Our \textsc{STARE} model consistently outperforms all baselines. For choice prediction using fixation data, it achieves the highest accuracy of 0.628, improving over the 2nd-best baseline, Chronos, by 5.19\%. Similar improvements are observed for AUC-ROC and F1. For item count prediction, \textsc{STARE} achieves the lowest RMSE (0.761), MAE (0.756), and MAPE (24.49\%) outperforming RETINA by 7.88\% and 8.36\%, and 8.34\% respectively. These results underscore the advantages of combining semantic tokenization and fixation-aware attention mechanisms in modeling human visual behavior.

We further evaluated model performance on the raw eye movement laptop dataset \citep{shi2013information}, which includes four input channels representing the x- and y-coordinates of the gaze from both the right and left eyes. This dataset focuses solely on the choice prediction task. As shown in Table~\ref{tab:benchmark_stare}, \textsc{STARE} achieves the best results across all metrics, reaching an accuracy of 0.798 compared to 0.783 for Chronos (a 1.92\% improvement), 0.789 for RETINA, and 0.781 for BERT. Improvements are similarly observed in AUC-ROC and F1, confirming that \textsc{STARE} generalizes well across both fixation-based and raw gaze data settings.

\subsection{Ablation Study of Tokenization and Attention Mechanisms}

\begin{table*}[t]
    \centering
    \scriptsize
    \renewcommand{\arraystretch}{1.3}
    \setlength{\tabcolsep}{4pt}
    \begin{tabular}{|l|ccc|ccc|ccc|}
        \hline
        & \multicolumn{3}{c|}{\textbf{Choice Prediction (Fixation)}} 
        & \multicolumn{3}{c|}{\textbf{Item Count Prediction (Fixation)}} 
        & \multicolumn{3}{c|}{\textbf{Choice Prediction (Raw Gaze)}} \\
        \cline{2-10}
        \textbf{Ablation Variant} 
        & \textbf{Accuracy} & \textbf{AUC-ROC} & \textbf{F1} 
        & \textbf{RMSE} & \textbf{MAE} & \textbf{MAPE (\%)} 
        & \textbf{Accuracy} & \textbf{AUC-ROC} & \textbf{F1} \\
        \hline
        RawSeq (no tokenization, no attention) 
        & 0.578 & 0.573 & 0.570 
        & 0.870 & 0.869 & 28.15 
        & 0.775 & 0.748 & 0.769 \\
        
        TokenOnly (Chronos tokenization, no attention) (\textbf{Chronos}) 
        & 0.597 & 0.591 & 0.589 
        & 0.817 & 0.819 & 26.53 
        & 0.783 & 0.758 & 0.772 \\
        
        TokenROI (ROI tokenization, no attention) 
        & 0.606 & 0.603 & 0.601 
        & 0.781 & 0.783 & 25.36 
        & 0.786 & 0.759 & 0.775 \\
        
        TokenROI + x- to y-axis CrossAttn 
        & 0.612 & 0.609 & 0.607 
        & 0.775 & 0.771 & 24.99 
        & 0.792 & 0.771 & 0.785 \\
        
        TokenROI + x- and y-axis CoAttn 
        & \underline{0.621} & \underline{0.614} & \underline{0.611} 
        & \underline{0.766} & \underline{0.760} & \underline{24.62} 
        & \underline{0.796} & \underline{0.775} & \underline{0.788} \\
        
        TokenROI + x- and y-axis Cross+CoAttn (\textbf{STARE}) 
        & \textbf{0.628} & \textbf{0.616} & \textbf{0.615} 
        & \textbf{0.761} & \textbf{0.756} & \textbf{24.49} 
        & \textbf{0.798} & \textbf{0.778} & \textbf{0.790} \\
        
        \hline
        \textbf{$\Delta$ vs. Best Baseline (\%)} 
        & \textbf{8.66\%} & \textbf{7.51\%} & \textbf{7.89\%} 
        & \textbf{6.89\%} & \textbf{7.65\%} & \textbf{12.99\%} 
        & \textbf{1.91\%} & \textbf{2.64\%} & \textbf{2.33\%} \\
        \hline
    \end{tabular}
    \caption{Ablation study results comparing STARE variants on fixation-based (PRS-in-Vivo) and raw gaze (laptop) eye-tracking data. Fixation-based evaluations include both choice prediction and item count prediction (RMSE, MAE, MAPE), while raw gaze results include only choice prediction. Second-best results are underlined.}
    \label{tab:ablation_stare}
\end{table*}

Our \textsc{STARE} model consists of two main components: an ROI-based tokenization module and an attention mechanism designed to model dependencies across fixation streams. The attention component includes two variants: cross-attention and co-attention, which can be used separately or in combination. To investigate the effect of each component on prediction performance using the two datasets, we conducted experiments across both tasks: choice prediction and item count prediction. Table~\ref{tab:ablation_stare} presents a detailed ablation study designed to isolate and quantify the individual and combined contributions of STARE’s core architectural elements: ROI-based tokenization, cross-attention, and co-attention. This analysis enables a fine-grained evaluation of how each component impacts the model's overall effectiveness across both classification and regression tasks.

We start with a baseline model that excludes both tokenization and attention, feeding into a shallow encoder either raw $(x, y)$ fixation coordinates (in the case of fixation data) or four input channels consisting of the x-and y-coordinates from the right and left eyes (in the case of raw gaze data), and passing it through the pre-rained T5 encoder \textsc{T5Embed} to generate contextualized embeddings. We denote this model as \textsc{RawSeq}. Next, we add the Chronos tokenization module without any attention to obtain \textsc{TokenOnly} and also add the proposed ROI tokenization - \textsc{TokenROI}. We then incorporate the attention module incrementally: \textsc{+CrossAttn} applies cross-attention between the $x$- and $y$-axis embeddings, \textsc{+CoAttn} applies symmetric co-attention, and the full model \textsc{STARE} jointly uses both mechanisms (\textsc{+Cross+CoAttn}).

\textbf{Fixation-based results.} For choice prediction, the \textsc{RawSeq} model, which lacks both tokenization and attention, already outperforms baselines (Bi-LSTM, the BERT models, and RETINA; see Table~\ref{tab:benchmark_stare}) across all classification metrics. For item count prediction, \textsc{RawSeq} also achieves lower RMSE, MAE and MAPE than the Bi-LSTM and BERT models. Adding Chronos' standard tokenization improves performance across both tasks, and the proposed ROI-based tokenization (\textsc{TokenROI}) yields further gains. For instance, ROI tokenization improves accuracy by 1.6\% and reduces RMSE by 4.4\% relative to Chronos. These results suggest that imposing semantic structure on gaze patterns via ROIs enhances fixation encoding for both binary and count-based predictions. Notably, prediction accuracy rises above 0.6 once ROI tokenization is introduced (e.g., 0.606 with \textsc{TokenROI}), even before applying attention.

Adding attention mechanisms further enhances performance across all metrics. For choice prediction, cross-attention increases accuracy by 1.0\% over \textsc{TokenROI}, while co-attention yields a larger gain of 2.5\%. Similarly, for item count prediction, cross-attention improves RMSE by 0.8\% and co-attention by 1.9\%, with parallel improvements in MAPE: from 25.36\% (TokenROI) to 24.99\% (CrossAttn) and 24.62\% (CoAttn). These results confirm the advantage of modeling interaction patterns between the horizontal and vertical fixation streams. The full \textsc{STARE} model, combining both attention types, achieves the best results across both tasks, reaching 0.628 accuracy, 0.761 RMSE, and 24.49\% MAPE- marking improvements of 5.2\% (accuracy) and 6.9\% (RMSE), and 7.7\% (MAPE) over the Chronos tokenization baseline without attention. These findings suggest that while horizontal and vertical eye movements are separately controlled, they are mutually influential. The larger gains from co-attention support the idea of bilateral coordination, consistent with prior neurological evidence \citep{horn2021functional, sparks2002brainstem}. Meanwhile, the added benefit of incorporating cross-attention suggests a smaller directional influence, particularly from horizontal to vertical movement during visual decision-making.

\textbf{Raw gaze results. } On the laptop dataset, which contains raw eye movement data and focuses solely on choice prediction, the ablation results show a similar trend. The \textsc{RawSeq} model (without tokenization or attention) already outperforms all baselines (see Table~\ref{tab:benchmark_stare}). Chronos tokenization (\textsc{TokenOnly}) increases accuracy to 0.783, and ROI tokenization (\textsc{TokenROI}) improves it further to 0.786. Adding cross-attention raises accuracy to 0.792 (a 0.8\% gain over \textsc{TokenROI}), and co-attention provides a larger gain to 0.796 (1.3\%). The full \textsc{STARE} model achieves the highest accuracy of 0.798, representing a 1.9\% improvement over Chronos. Similar improvements are observed in AUC-ROC and F1 metrics. These results highlight the importance of attention-based modeling of spatial gaze dynamics. In particular, co-attention between the left and right eyes appears especially beneficial, consistent with the notion of separate neural control for each eye. The additional improvement from cross-attention suggests a (smaller) directional signal, likely driven by right-eye dominance, which contributes to improved predictions on raw gaze sequences.

The ablation experiments thus demonstrate the predictive performance of STARE, and highlight the contributions of its components across both fixation-based and raw gaze data. It suggests that horizontal and vertical eye movements, as well as right and left eye trajectories, are separately controlled, influencing each other mutually, which is commensurate with their separate neural control \citep{horn2021functional, sparks2002brainstem}. The performance gains achieved by combining cross- and co-attention mechanisms support the idea of bilateral coordination alongside a smaller directional influence, particularly from horizontal and right-eye movements. These findings point to the importance of jointly modeling spatial gaze dynamics when predicting visual decision-making.

\subsection{Partial Time and Data to Predict}
\label{subsec:partialData}

\begin{figure*}
  \centering
  \includegraphics[width=0.6\linewidth]{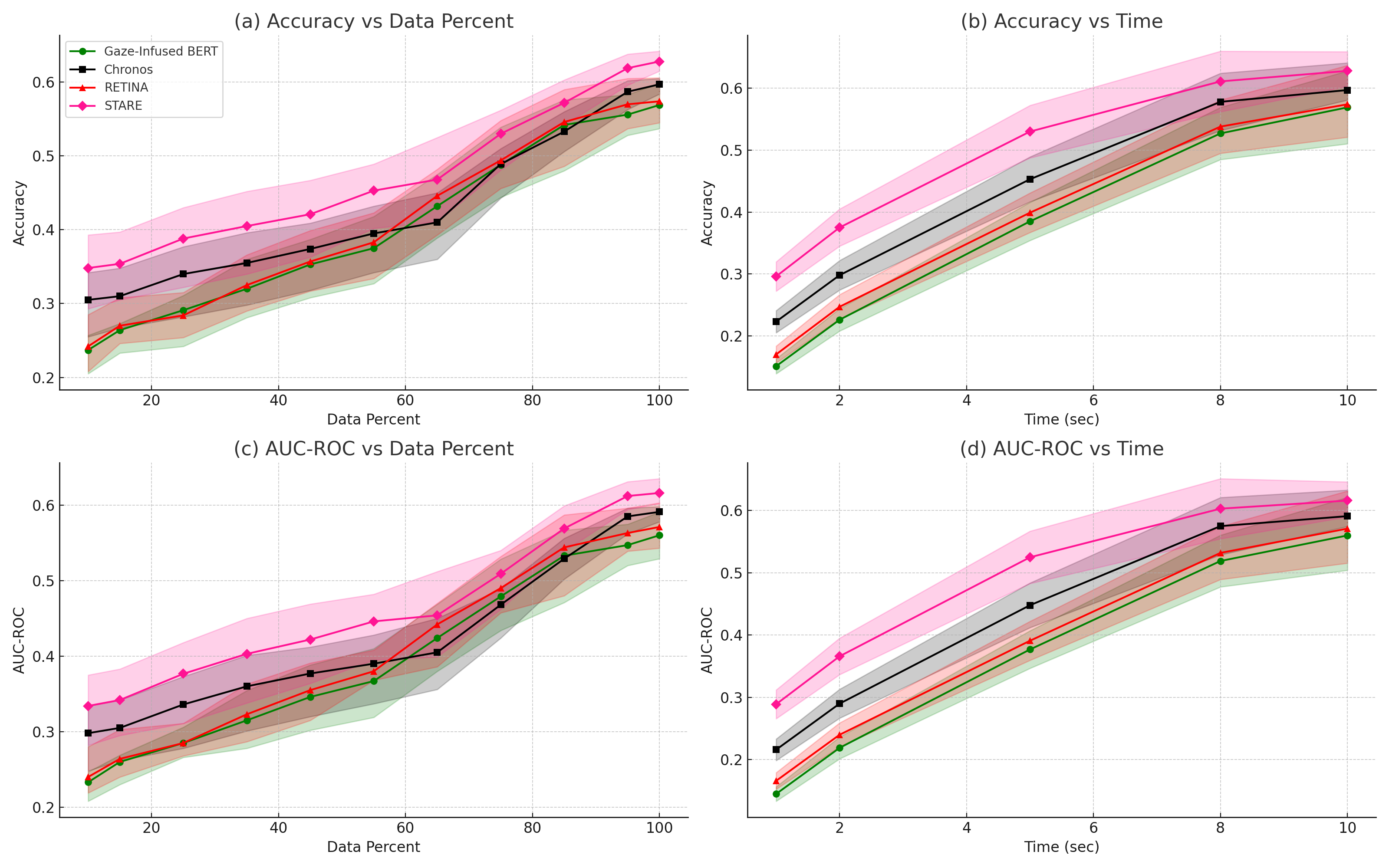}
  \caption{Gaze-Infused BERT, RETINA, Chronos and STARE prediction performance (accuracy and AUC-ROC) using partial data (Figures a and c) and partial time (Figures b and d)}
  \label{fig:dataPredictSettings}
\end{figure*}

An important research question in the eye-tracking literature \citep{martinovici2023attention, shimojo2003gaze,byrne2023predicting, unger2024predicting} is \emph{when} the prediction of consumers' choice based on eye movements begins to work well, and \emph{how much} data is needed to obtain satisfactory performance. To investigate this, we trained the four most competitive prediction models, STARE, Chronos, RETINA, and Gaze-Infused BERT under two complementary experimental conditions on the PRS-in-Vivo fixation data. We use that dataset to illustrate these predictions because fixation data is by far the most common type of eye movement data used in practice. We use subsequences of the fixation data, starting from the onset of the recording, to predict pick-any decision making, while varying the length of the subsequences either by percentage of data or by absolute time.

In the first condition, termed "partial data to predict," we use for each participant the first 10\% to 100\% of the eye fixations, in steps of 5\%, to predict their final product choices. Since the total decision time varies by participant, the same relative percentage of fixations corresponds to different amounts of absolute time. In the second condition, "partial time to predict," we therefore use the eye-fixation data from the first 1, 2, 5, 8, and 10 seconds for each participant. Unlike relative percentage sampling, this condition ensures equal time windows across participants, making it more relevant for real-time prediction settings.

Figure~\ref{fig:dataPredictSettings} shows the accuracy and AUC-ROC performance across the two conditions using the PRS-in-Vivo retail fixation data. For each model, we report the mean accuracy and AUC-ROC across participants, along with the range (minimum and maximum) of values obtained.

As shown in Figure~\ref{fig:dataPredictSettings}(a), prediction accuracy generally improves with increasing data availability, as expected. Notably, both \textsc{STARE} and Chronos exhibit relatively strong performance even with minimal input, achieving 0.348 and 0.305 accuracy, respectively, when only 10\% of the data is used for training. 
In contrast, RETINA and Gaze-Infused BERT improve more gradually as more data becomes available. A distinct inflection point in performance appears around 65\% of the data, particularly for the transformer-based models Chronos (0.410 accuracy) and STARE (0.468 accuracy). This reflects the models’ need for a critical mass of fixation information to reach higher predictive accuracy. After this threshold, accuracy increases more steeply, with \textsc{STARE} ultimately reaching the highest final accuracy (0.628), followed by Chronos (0.597), RETINA (0.574), and Gaze-Infused BERT (0.569). AUC-ROC trends in Figure~\ref{fig:dataPredictSettings}(c) are consistent with the accuracy results, reinforcing the comparative advantage of STARE and Chronos across data fractions. In terms of both accuracy and AUC, STARE's performance is uniformly better than that of all other models across the entire range of percentage of data used.

Turning to the ``partial time to predict" condition, Figure~\ref{fig:dataPredictSettings}(b) shows that prediction accuracy improves rapidly in the first few seconds of viewing. After 5 seconds STARE reaches an accuracy of 0.530 and Chronos achieves 0.453, while performance increases begin to taper off somewhat. After 10 seconds of data, which is less than the decision time of most participants, accuracy reaches ceiling levels at 0.628 for STARE, 0.597 for Chronos, 0.574 for RETINA, and 0.569 for Gaze-Infused BERT (see Table~\ref{tab:benchmark_stare}). This represents a relative improvement of 18.5\% for \textsc{STARE} and 17.4\% for Chronos between the 5- and 10-second marks. As Figure~\ref{fig:dataPredictSettings}(b) demonstrates, STARE leads consistently across all time windows for the accuracy measure, and Chronos comes in second. Figure~\ref{fig:dataPredictSettings}(d) confirms that AUC-ROC exhibits a similar trend, with a clear separation between the top (STARE/Chronos) and bottom (RETINA/Gaze-Infused BERT) models, with STARE again outperforming the other models across the entire range of time used to predict choice.

In summary, STARE achieves the best performance in all cases under both relative and absolute time slicing, as demonstrated in Figures~\ref{fig:dataPredictSettings} (a)-(d). Its robustness across participants and time horizons suggests that it is well-suited for early-stage prediction of and intervention in the choice process in real-time applications. In contrast, while Chronos, RETINA and Gaze-Infused BERT show solid performance, they are consistently outpaced when data is limited. This experiment confirms that the STARE transformer-based time-series model can predict choices reliably even in early stages of decision making, presumably before participants have reached their choice decisions. It thus further supports the importance of temporal modeling, in particular via the STARE model architecture, tokenization and co- and cross-attention mechanisms.

\section{Conclusion}

We proposed a novel architecture, called STARE, for the prediction of consumer choice from eye-tracking data, building on the Chronos model.
The STARE model uses a novel spatio-temporal tokenization, based on semantically meaningful Regions-of-Interest (ROI) in the decision environment and co- and cross-attention mechanisms on top of a pre-trained T5 embedding, that capture the dependencies in the spatio-temporal input channels. The model accommodates horizontal and vertical movements captured by both (unequally spaced) fixation data, as well as (equally spaced) raw data on movements of the left and right eyes. Ablation studies show that the tokenization strategy, as well as the attention mechanisms, capture features of the eye movements that contribute to better predictions of consumer choice. STARE outperforms state-of-the-art benchmark models, such as Chronos, RETINA and Gaze-Infused BERT, in predicting choice both from the complete data and from partial eye-movement data selected by percentage of data or amount of time, thus predicting individuals' choices accurately within five to ten seconds, \emph{well before} a decision is reached. 

STARE offers the potential to connect \textit{in silico} attention mechanisms with \textit{in vivo} neurophysiologic and attention processes underlying human eye movements. Specifically, the co-and cross-attention mechanisms inform us about the human attention processes underlying the coordination of respectively horizontal and vertical (for fixation data) and left-and right eye movements (for raw gaze data), thus offering a first step towards connecting Deep Learning attention architectures to human attention processes, and enhancing our understanding of how gaze behavior influences decision-making. In addition to providing a foundational model for eye-tracking data, STARE may be applied to other spatio-temporal irregularly spaced multivariate time-series data, which we expect to be explored in future research. 

\appendices
\section*{Acknowledgment}
The authors would like to thank Hideo Maldonado, Oliver Blanchet and others at the PRS-in-Vivo team for their support and providing the data. The authors also acknowledge the contribution of Chen Xing in data processing.




%

\bibliographystyle{plain}
\bibliography{bare_jrnl}



%







\end{document}